\documentclass[letterpaper, 10 pt, conference]{ieeeconf}  % Comment this line out if you need a4paper

\IEEEoverridecommandlockouts                              % This command is only needed if 
\usepackage{graphicx}
\usepackage{url}  %IVM added this because of the URLs in the reference list

\usepackage{amsmath}

\title{\LARGE \bf
Remote Sensing Image Classification using Transfer Learning and Attention Based Deep Neural Network}

\author{Lam~Pham$^{1*}$, 
        Khoa~Tran$^{2*}$,
        Dat~Ngo$^{4}$,
        Jasmin~Lampert$^{1}$, 
        Alexander~Schindler$^{1}$, 
\thanks{L. Pham, J. Lampert, and A. Schindler are with Center for Digital Safety \& Security, Austrian Institute of Technology, Austria.}%
\thanks{K. Tran is with University of Science and Technology, University of Danang, Viet Nam.}%
\thanks{D. Ngo is with School of Computer Science and Electronic Engineering, University of Essex, UK.}%
\thanks{(*) Main and equal contribution into the paper.}
}

\begin{document}

\maketitle
\thispagestyle{empty}
\pagestyle{empty}

%%%%%%%%%%%%%%%%%%%%%%%%%%%%%%%%%%%%%%%%%%%%%%%%%%%%%%%%%%%%%%%%%%%%
\begin{abstract}
The task of remote sensing image scene classification (RSISC), which aims at classifying remote sensing images into groups of semantic categories based on their contents, has taken the important role in a wide range of applications such as urban planning, natural hazards detection, environment monitoring,vegetation mapping, or geospatial object detection.
During the past years, research community focusing on RSISC task has shown significant effort to publish diverse datasets as well as propose different approaches to deal with the RSISC challenges.
Recently, almost proposed RSISC systems base on deep learning models which prove powerful and outperform traditional approaches using image processing and machine learning.  
In this paper, we also leverage the power of deep learning technology, evaluate a variety of deep neural network architectures, indicate main factors affecting the performance of a RSISC system.
Given the comprehensive analysis, we propose a deep learning based framework for RSISC, which makes use of the transfer learning technique and multihead attention scheme.
The proposed deep learning framework is evaluated on the benchmark NWPU-RESISC45 dataset and achieves the best classification accuracy of 94.7\% which shows competitive to the state-of-the-art systems and potential for real-life applications.

\indent \textit{Relevant Items}--- Convolutional Neural Network (CNN), Transfer Learning, Attention, Remote Sensing Image, Data Augmentation.
 
\end{abstract}
%%%%%%%%%%%%%%%%%%%%%%%%%%%%%%%%%%%%%%%%%%%%%%%%%%%%%%%%%%%%%%%%%%%%%%%%%%%%%%%%
\section{INTRODUCTION}
Remote sensing image scene classification (RSISC), which was early mentioned in~\cite{early_rsic, old_01}, has recently attracted much research attention as leveraging the power of deep learning techniques.  
Indeed, while handcrafted features using image processing techniques such as color histogram~\cite{early_rsic, aid_dataset}, Gabor-based texture features~\cite{early_rsic},  SIFT description~\cite{old_01, aid_dataset} and traditional machine learning models for classification such as Support Vector Machine (SVM)~\cite{early_rsic, old_01}, Gaussian Mixture based Clustering or Classification~\cite{old_02, aid_dataset} were widely applied before, recent publications for RSISC have presented a wide range of deep learning models, mainly basing on deep convolutional neural network architectures.
Generally, to aggregate global features in a remote sensing image, recently proposed models make use of the convolution operation, stack multiple convolutional layers to make the networks deeper and easier to train.
For instances, ResNet50 and Resnet101 architectures were explored in~\cite{sota_11}, presenting very competitive results on different benchmark datasets of NWPU-RESISC45~\cite{nwpu_dataset}, AID~\cite{aid_dataset}, UC-Merced~\cite{early_rsic}, and WHU-RS19~\cite{whu_rs19_dataset}.
Similarly, authors in~\cite{sota_05} deployed EffectiveNet based models while VGG based network architectures such as AlexNet, VGG-16, or GoogLeNet were fine-tuned in~\cite{vgg_finetune}.

It can be seen that recent RSISC systems make effort to explore or fine-tune specific neural network architectures, but have not provided a detailed analysis of factors affecting the network performances or a comprehensive comparison among different network architectures.
This inspires us to conduct an extensive experiment in this paper and present three main contributions: 
\begin{itemize}
\item We evaluate how main factors in a deep neural based framework such as data augmentation, neural network architecture, training strategies affect RSISC performance.
Given the comprehensive analysis, we propose an effective deep learning framework for RSISC task.
\item We further enhance RSISC performance by proposing a multihead attention based network layer which is used to replace traditional global average or max pooling layers.
The proposed multihead attention based layer is developed to deeply learn feature map across the width and height dimensions, then generate more distinct features rather than those extracted from a traditional global average or max pooling layer. 
\item We evaluate various deep neural networks, indicate the most effective network architectures for RSISC task. We then evaluate whether applying an ensemble method of individual high-performance models can help to improve the performance.
\end{itemize}

\section{Proposed deep learning frameworks}
\label{framework}

\begin{figure}[t]
    	\vspace{-0.2cm}
    \centering
    \includegraphics[width =1.0\linewidth]{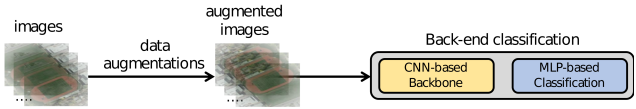}
    	\vspace{-0.5cm}
	\caption{The high-level architecture of proposed deep learning frameworks.}
    \label{fig:A1}
\end{figure}

As Figure~\ref{fig:A1} shows, the high-level architecture of our proposed deep learning frameworks comprise two main steps: (1) the data augmentation to make the input images diverse and (2) the back-end deep neural networks for classification.

\subsection{Data augmentation}
\label{augmentation}
In this paper, we apply four methods of data augmentation: Image Rotation (IR)~\cite{rotation_aug}, Random Cropping (RC)~\cite{rotation_aug}, Random Erasing (RE)~\cite{spec_crop}, and Mixup (Mi)~\cite{mixup1, mixup2}.
In particular, all images in the target dataset are rotated with angles set to 90, 180, and 270, referred to as Image Rotation (IR).
By using this method, the dataset increases four times.
Then, the images are randomly grouped in batches with batch size set to 32.
For each batch, the images are randomly cropped with a reduction of 10 pixels on both width and height dimensions, referred to as Random Cropping (RC).
Then, random 20 pixels on both width and height dimensions of each image are erased, referred to as Random Erasing (RE).
Finally, the images are mixed together with random ratios, referred to as Mixup (Mi).
As we use both Uniform or Beta distributions to generate the mixup ratios, the batch size increases three times from 32 to 96.
Batches of 96 images are then fed into the deep neural networks for classification.
As Random Cropping (RC)~\cite{rotation_aug}, Random Erasing (RE)~\cite{spec_crop}, and Mixup (Mi)~\cite{mixup1, mixup2} are used on batches of images during the training process, they are grouped in to the on-line data augmentation. 
Meanwhile, Image Rotation (IR)~\cite{rotation_aug} is referred to as the off-line data augmentation as this method is applied on all training dataset before the training process.
%++++++++++++++
%++++++++++++++
%--------------------------------------------
\begin{table}[t]
    \caption{The proposed MLP-based classification} 
        	\vspace{-0.2cm}
    \centering
    \scalebox{1.0}{
    \begin{tabular}{|l |c |} 
        \hline 
            \textbf{Setting Layers}   &  \textbf{Outputs}  \\
       \hline          
                     FC(4096) - ReLU - Dr(0.2)   &  4096  \\
                     FC($C$) - Softmax   &  $C$  \\

       \hline 
           %	\vspace{-1cm}
    \end{tabular}
    }
        \vspace{-0.3cm}
    \label{table:mlp} 
\end{table}
\subsection{Proposed deep neural networks for classification}
\label{neural}

As the high-level architecture of the proposed deep learning frameworks is shown in Figure~\ref{fig:A1}, the back-end deep neural network, which presents a convolutional-based architecture, can be separated into two main parts: (1) The convolutional-based backbone (CNN-based backbone), which is performed by trunks of convolutional-based layers, helps to transfer the input images to condensed feature map; and (2) the multiplayer perceptron based classification (MLP-based classification) for classifying the feature map extracted from CNN-based backbone into certain categories.

In this paper, the proposed MLP-based classification is performed by fully connected layer (FC(channel number)), rectify linear unit (ReLU)~\cite{relu}, Dropout (Dr(drop ratio))~\cite{dropout}, and Softmax layer, as described in Table~\ref{table:mlp}.
Notably, the number of channel at the second fully connected layer is set to $C$ that presents the number of categories of the target RICSC  dataset (i.e., for an example, $C$ is set to 45 for NWPU-RESISC45 dataset~\cite{nwpu_dataset}).
Meanwhile, to achieve the best architecture for the CNN-based backbone, we reuse and evaluate a wide range of benchmark deep neural networks.
In particular, we firstly evaluate residual based network architectures of ResNet152, ResNet152V2, DenseNet121, DenseNet201based, which present a deep trunk of convolutional layers, showing a scale up of convolutional layers by depth.
We then evaluate InceptionV3 architecture which scales up convolutional layers by width with multiple kernels.
We also evaluate InceptionResNetV2 network which shows a hybrid form between residual based and inception based architectures.
Recently, EfficientNet architectures have developed to aim at uniformly scaling all dimensions of depth, width, and image resolution~\cite{effnet}.
We therefore evaluate two forms of EfficientNet architectures of EfficientNetB0 and EfficientNetB4 in this paper.
These benchmark networks of ResNet152, ResNet152V2, InceptionV3, InceptionResNetV2, DenseNet121, DenseNet201, EfficientNetB0, EfficientNetB4 are available in the Keras library~\cite{keras_app}.

\begin{figure}[t]
    	\vspace{-0.2cm}
    \centering
    \includegraphics[width =1.0\linewidth]{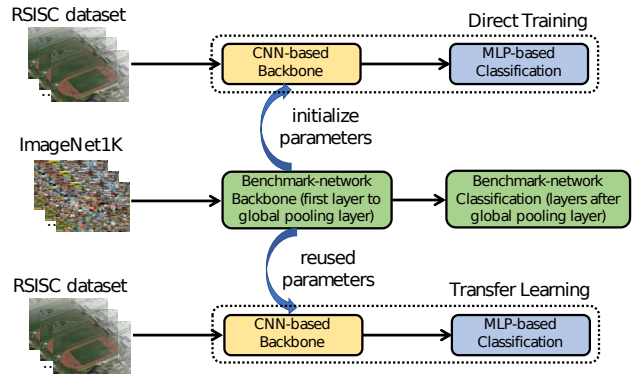}
    	\vspace{-0.5cm}
	\caption{Direct training and transfer learning strategies by reusing the network backbone from benchmark deep neural network architectures.}
    \label{fig:A3}
\end{figure}

By reusing these benchmark networks, we propose two training strategies: direct training and transfer learning as shown in Figure~\ref{fig:A3}.
In the direct training strategy as shown in the upper part of Figure~\ref{fig:A3}, we reuse the first layer to the global pooling layer of these benchmark networks, referred to as the benchmark-network backbone, to perform the proposed CNN-based backbone.
All trainable parameters of the proposed deep neural networks in the direct training strategy (i.e., all trainable parameters of both the CNN-based backbone and the MLP-based classification) are initialized with random values of mean 0 and variance 0.1.
In the transfer learning strategy as shown in the lower part in Figure~\ref{fig:A3}, the benchmark network architectures were trained with the large-scale ImageNet1K dataset~\cite{imagenet_dataset} in advance.
Then, trainable parameters of the benchmark-network backbone are transferred to the CNN-based backbone.
In other words, only trainable parameters of the MLP-based classification are initialized while those of the CNN-based backbone are reused from the pre-trained benchmark network architectures in the transfer learning strategy.
While the direct training strategy uses the learning rate of 0.0001, a lower learning rate of 0.00001 is set in the transfer learning strategy.

\label{framework}
%++++++++++++++
%++++++++++++++
%++++++++++++++
\begin{figure*}[t]
    	\vspace{-0.2cm}
    \centering
    \includegraphics[width =1.0\linewidth]{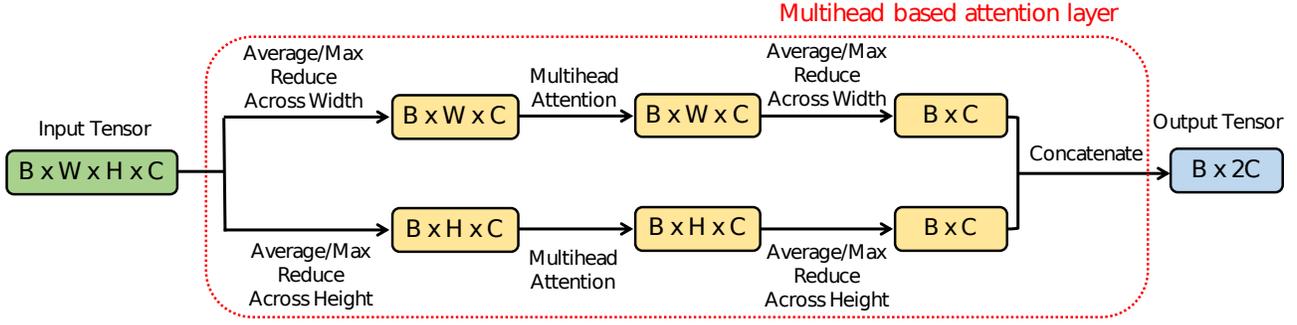}
    	\vspace{-0.5cm}
	\caption{The proposed multihead attention based layer architecture to replace the global average or max pooling layer in CNN-based backbone (B: Batch dimension; W: Width dimension; H: Height dimension; C: Channel dimension)}
    \label{fig:A2}
\end{figure*}
%++++++++++++++

\subsection{Propose a multihead attention based network layer}
\label{attention}

Inspire that traditional global pooling layers may not perform well as a global max or average value across the width and height dimensions cannot represent for an entire two-dimensional feature map, we therefore evaluate whether applying the attention technique can help to enhance the performance.
In particular, we replace the global pooling layer in the CNN-based backbone by a multihead attention based layer as shown in Figure~\ref{fig:A2}.
The proposed multihead attention based layer is developed to aim at learning distinct features across the width and height dimensions using multihead attention scheme~\cite{multihead_att}.
To this end, let us consider the shape of an input tensor of the proposed multihead attention based layer as $[B\times W\times H\times C]$, where $B$, $W$, $H$, and $C$ are the batch size, width dimension, height dimension, and channel dimension, respectively.
We then reduce either the width or the heigh dimension of the input tensor using average or max pooling layers, generating tensors of $[B\times  W\times C]$ or $[B\times H\times C]$ with two independent data streams.
Next, the multihead attention scheme~\cite{multihead_att} is applied on two data streams, each of which focuses on learning distinct features across either the width dimension (i.e., the upper data stream in Figure~\ref{fig:A2}) or the height dimension (i.e., the lower data stream in Figure~\ref{fig:A2}).
Notably, we empirically set the number of heads to 16, set the key dimension to 128, and retain the shape of input tensors.
Then, average or max pooling layers are again applied to reduce the width or height dimensions, achieving tensors of  $[B\times C]$ on each data stream.
Finally, two data streams are concatenated to generate a tensor of $[2B\times C]$ before transferring into MLP-based classification.

\subsection{Apply an ensemble to enhance the performance}
\label{ensemble}

Inspire that predicted probabilities obtained from individual deep neural network architectures can complement each other and a fusion of these predicted probabilities can help to improve the performance~\cite{hydra_ens, pham2021audio, pham2022deep}, we therefore propose an ensemble of predicted probabilities in this paper, referred to as PROD late fusion.
Let us consider predicted probability of each deep neural network as  \(\mathbf{\bar{p_{n}}}= (\bar{p}_{n1}, \bar{p}_{n2}, ..., \bar{p}_{nC})\), where $C$ is the category number and the \(n^{th}\) out of \(N\) networks evaluated, the predicted probability after PROD fusion \(\mathbf{p_{prod}} = (\bar{p}_{1}, \bar{p}_{2}, ..., \bar{p}_{C}) \) is obtained by:
\begin{equation}
\label{eq:mix_up_x1}
\bar{p_{c}} = \frac{1}{N} \prod_{n=1}^{N} \bar{p}_{nc} ~~~  for  ~~ 1 \leq n \leq N 
\end{equation}
Finally, the predicted label  \(\hat{y}\) is determined by 
\begin{equation}
    \label{eq:label_determine}
    \hat{y} = arg max (\bar{p}_{1}, \bar{p}_{2}, ...,\bar{p}_{C} )
\end{equation}

\section{Experiments and results}

\subsection{Dataset and metric}
\label{dataset}
In this paper, we evaluate the benchmark NWPU-RESISC45 dataset~\cite{nwpu_dataset} which consists of 31500 remote sensing images collected from more than 100 countries and regions all over the world.
31500 images are separated into 45 scene classes, each of which comprises 700 images in RGB color format with a size of 256$\times$256 pixels. 
%The spatial resolution varies from about 30 to 0.2 m per pixel for most of the scene classes with an exception of island, lake, mountain, and snow-berg showing lower spatial resolutions. 
Compare to the other benchmark datasets of AID~\cite{aid_dataset}, UC-Merced~\cite{early_rsic}, and WHU-RS19~\cite{whu_rs19_dataset}, NWPU-RESISC45 dataset presents the larger number of categories and the larger number of images per each category.

To compare with the state-of-the-art systems, we follow the original settings in~\cite{nwpu_dataset}, then split the NWPU-RESISC45 dataset into Training-Testing subsets with two different ratios: 20\%-80\%  and 10\%-90\%.
We also obey the original paper~\cite{nwpu_dataset} using Accuracy (Acc.\%) as the metric to evaluate our proposed systems in this paper.

\subsection{Experimental settings}
\label{setting}
Due to use of the Mixup~\cite{mixup2} data augmentation, labels are no longer in one-hot encoding format. Therefore, we use a Kullback-Leibler divergence (KL) loss~\cite{kl_loss} shown in Eq. (\ref{eq:kl_loss}) below.
\begin{align}
   \label{eq:kl_loss}
   Loss_{KL}(\Theta) = \sum_{n=1}^{N}\mathbf{y}_{n}\log \left\{ \frac{\mathbf{y}_{n}}{\mathbf{\hat{y}}_{n}} \right\}  +  \frac{\lambda}{2}||\Theta||_{2}^{2}
\end{align}
where  \(\Theta\) are trainable parameters, constant \(\lambda\) is set initially to $0.0001$, batch size $N$, $\mathbf{y_{i}}$ and $\mathbf{\hat{y}_{i}}$  denote expected and predicted results.
All deep learning networks proposed are constructed with Tensorflow, and the experiments were performed on two 24GB Nvidia Titan-GPUs for both training and inference jobs.

%--------------------
\begin{table*}[t]
    \caption{Performance (Acc.(\%)/Training Epoches) comparison among deep learning frameworks with direct training (DR) and transfer learning (TL), with or without using data augmentation (Aug.) and attention (Att.) on the benchmark NWPU-RESISC45 dataset with 20\% training setting.} 
        	\vspace{-0.2cm}
    \centering
    \scalebox{0.9}{

    \begin{tabular}{|c   |c|c|c|c  |c|c|c|c|}  
        \hline 
	     \textbf{Configuration/Networks}      &\textbf{ResNet152}   &\textbf{ResNet152V2}  &\textbf{DenseNet121} &\textbf{DenseNet201} &\textbf{InceptionV3} &\textbf{InceptionResNetV2} &\textbf{EfficientNetB0} &\textbf{EfficientNetB4}\\
	          
        \hline 
       
        \textbf{DT}            &56.2/144      &46.0/102   &56.7/83     &57.5/102     &45.5/146   &64.7/58    &49.8/132   &43.5/131  \\                         
        \textbf{TL}            &86.1/32       &78.6/34    &88.7/65     &88.1/66      &85.6/66    &84.4/31    &86.2/76    &81.5/49   \\                     
            %\hline
        \textbf{TL \& Aug.}    &89.1/15       &91.1/16    &91.8/31     &91.7/23      &89.6/18    &87.0/16    &91.3/38    &91.8/22   \\
        \textbf{TL \& Aug. \& Att.}   &89.3/10       &91.7/10    &\textbf{92.8/13}  &\textbf{92.9/8}    &90.9/17   &90.5/11    &\textbf{92.9/26}   &\textbf{92.8/12}   \\  
                \hline 

      \end{tabular}
    }
    %\vspace{-0.3cm}
    \label{table:res_01} 
\end{table*}
%-------------------
%--------------------
\begin{table*}[t]
    \caption{Performance comparison using late fusion of DenseNet201 and EfficientNetB4, max and average pooling layers inside the proposed attention on the benchmark NWPU-RESISC45 dataset with 20\% training setting.} 
        	\vspace{-0.2cm}
    \centering
    \scalebox{1.0}{

    \begin{tabular}{|c|c|c|c|c|}  
        \hline     
                \textbf{Pooling layers} &Max + Average    &Max + Average      &Average                       &Max + Average \\
        \hline 
	            \textbf{Networks}       &DenseNet201      &EfficientNetB4     &DenseNet201 + EfficientNetB4  &DenseNet201 + EfficientNetB4   \\	          
        \hline 
                \textbf{Acc.(\%)} &93.6 &93.1    &94.3   &94.7 \\
        \hline 

      \end{tabular}
    }
    %\vspace{-0.3cm}
    \label{table:res_02} 
\end{table*}
%-------------------

\subsection{Performance comparison between direct training and transfer learning approaches with and without using data augmentation methods and the proposed attention}
We firstly compare the direct training (DT) and transfer learning (TL) strategies without using both data augmentation methods and the proposed attention layer.
As results are shown in Table~\ref{table:res_01}, we can see that the networks with the transfer learning strategy significantly outperform those using the direct training.
Almost proposed deep learning networks using the transfer learning strategy achieve more than 80\% with an exception of ResNet152V2.
The transfer learning strategy also helps to reduce the training time, especially for the large networks of ResNet152, ResNet152V2, or EfficientNetB4.
Compare among deep neural networks, DenseNet based architectures outperform the others regarding using the transfer learning strategy, recording classification accuracy scores of 88.7\% and 88.1\% for DenseNet121 and DenseNet201, respectively.

When we apply the data augmentation methods, the performance of EfficientNetB4 significantly improves by 10\% while the other networks show an average improvement of 3\%.
By using the transfer learning strategy and data augmentation methods, the EfficinetNet and DenseNet based architectures are competitive and outperform ResNet and Inception based networks.
Although training times (i.e, the epoch numbers) significantly reduces, each training epoch needs more time to finish as the data augmentation methods increase the number of input data.

Next, we evaluate the transfer learning strategy using both data augmentation methods and the proposed attention layer (i.e., notably, the average pooling layer is used in the proposed attention layer).
As the final line in Table~\ref{table:res_01} shows, InceptionV3 and DenseNet, EfficientNet based architectures improve by an average score of 1\%.
Meanwhile, an improvement of more than 3\% is presented with InceptionResNetV2, but a slightly improvement is for ResNet based deep neural networks.
Compare performances among networks, DenseNet and EfficientNet based architectures are competitive and outperform the others, recording an average score of 92.8\%.
Applying attention layer also helps to reduce training time, especially for the large networks of DenseNet201 and EfficientNetB4 with 8 and 12 training epoches.

As DenseNet201 and EfficientNetB4 achieve the best performances compared to the other architectures, we conduct PROD late fusion of the predicted probabilities obtained from these networks as mentioned in Section~\ref{ensemble}.
Additionally, as the average or max pooling layer used in the proposed attention focuses on either max or average of feature map across the width or height dimension, we evaluate whether the average and max pooling layers can complement.
In particular, RPOD late fusion of predicted probabilities from DenseNet201 or EfficientNetB4 using either max or average pooling layer are evaluated. 
As the ensemble results are shown in Table~\ref{table:res_02}, we can see that ensembles of max and average pooling layers or ensembles of DenseNet201 and EfficientNetB4 network architectures can help to improve the performance.
When we conduct ensemble of four individual models: DenseNet201 with max pooling layer,  DenseNet201 with average pooling layer, EfficientNetB4 with max pooling layer, and EfficientNetB4 with average pooling layer, we can achieve the best classification accuracy score of 94.7\%. 

\subsection{Performance comparison to the state of the art}
%--------------------
\begin{table}[t]
    \caption{Performance (Acc.\%) comparison to the state-of-the-art systems on the benchmark NWPU-RESISC45 dataset with two splitting settings.} 
        	\vspace{-0.2cm}
    \centering
    \scalebox{1.0}{

    \begin{tabular}{|l|c|c|} 
        \hline 
        \textbf{Methods}  &\textbf{10\% training} &\textbf{20\% training} \\
	    \hline         
APDC-Net~\cite{sota_03}                     &85.9    &87.8 \\
D-CNN with GoogLeNet~\cite{sota_02}         &86.9    &90.5 \\  
VGG-16 + MTL~\cite{sota_11}                 &-       &91.5 \\ 
MG-CAP (Log-E)~\cite{sota_01}               &89.4    &91.7 \\ 
MG-CAP (Bilinear)~\cite{sota_01}            &89.4    &93.0 \\
EfficientNet-B0-aux~\cite{sota_05}          &90.0    &-  \\
MG-CAP (Sqrt-E)~\cite{sota_01}              &90.8    &93.0 \\ 
ResNeXt-50 + MTL~\cite{sota_11}             &-       &93.8 \\
ResNeXt-101 + MTL~\cite{sota_11}            &91.9    &94.2 \\
EfficientNet-B3-aux~\cite{sota_05}          &91.1    &-  \\
SE-MDPMNet~\cite{sota_10}                   &91.8    &94.1 \\
Xu’s method~\cite{sota_12}                  &91.9    &94.4 \\
%ransformer~\cite{sota_13}                  &93.1    &95.6 \\ 
        \hline 
  Our systems                               &92.6    &94.7 \\
         \hline 

      \end{tabular}
    }
    %\vspace{-0.3cm}
    \label{table:res_03} 
\end{table}
%-------------------

As Table~\ref{table:res_03} shows the performance comparison with the state-of-the-art systems, we can see that our best model are competitive, recording accuracy scores of 92.6\% and 94.7\% with training ratios of 10\% and 20\%, respectively.
%Our best results are only lower than those from the Transformer-based system proposed in~\cite{sota_13}. 

\section{Conclusion}
This paper has presented an exploration of various deep learning models for remote sensing image classification (RSISC).
By conducting extensive experiments, we indicate that applying multiple techniques of transfer learning, data augmentation, attention scheme, and ensemble to DenseNet and EfficientNet based deep learning frameworks is effective to achieve a high-performance RSISC system. 

\addtolength{\textheight}{-11cm}   % This command serves to balance the column lengths
                                  % on the last page of the document manually. It shortens
                                  % the textheight of the last page by a suitable amount.
                                  % This command does not take effect until the next page
                                  % so it should come on the page before the last. Make
                                  % sure that you do not shorten the textheight too much.

%\begin{thebibliography}{99}
\bibliographystyle{IEEEbib}
\bibliography{refs}
%\end{thebibliography}
\end{document}